\documentclass[letterpaper]{article}
\usepackage{natbib,alifeconf}

\title{Modeling the Evolution of Gene-Culture Divergence}
\author{Chris Marriott$^{1}$ \and Jobran Chebib$^2$ \\
\mbox{}\\
$^1$University of Washington, Tacoma, WA, USA 98402 \\
$^2$University of Z\"urich, Z\"urich, Switzerland 8057 \\
dr.chris.marriott@gmail.com}

\begin{document}
\maketitle

\begin{abstract}
We present a model for evolving agents using both genetic and cultural inheritance mechanisms.  Within each agent our model maintains two distinct information stores we call the genome and the memome.  Processes of adaptation are modeled as evolutionary processes at each level of adaptation (phylogenetic, ontogenetic, sociogenetic).  We review relevant competing models and we show how our model improves on previous attempts to model genetic and cultural evolutionary processes.  In particular we argue our model can achieve divergent gene-culture co-evolution. 
\end{abstract}

\section{Introduction}
Evolutionary computation, a field that exploits the power of evolution, is a powerful tool for optimization, creativity and the study of evolutionary forces.  \cite{h75}  helped popularize evolutionary computation by applying the principles of a basic evolutionary model to a computational task.  \cite{d76} also considered the minimal requirements of evolution and proposed a simple model.

Dawkins applied his model to the domain of human culture introducing the field of memetics (see also \cite{d95}).   Dawkins suggests that as separate domains the realm of genetics and the realm of memetics both follow the same evolutionary principles.  Theories of genetic and cultural co-evolution recognize that these two domains are not separate but parts of the same evolving system.  While Dawkins' simplified model addresses a single evolving system in isolation, we are interested in a model that incorporates the interactions between genes \textit{and} memes as both evolve. 

There have been many attempts to characterize the nature of cultural evolution coming from diverse motivations.  For instance the term memetic algorithms now refers to a field of combinatorial optimization.  \cite{m89}  defined this new model in this way:

\begin{quote}
Memetic algorithms is a marriage between a population-based global
search and the heuristic local search made by each of the individuals.
\end{quote}

\noindent While the focus of Moscato was combinatorial optimization, where this model has proven of value, he has not created a model incorporating both genetic and cultural evolution.  A model of cultural evolution must also incorporate an \emph{exchange of learned information between individuals}. In this sense Moscato's model falls short of our aims.  (It is worth noting that some implementations of memetic algorithms do incorporate an exchange of information.)

In developing our own model of genetic and cultural evolution \citep{MC14} we have considered the characteristics of an acceptable model.  In our opinion any acceptable model will incorporate three modes of adaptation: phylogeneitc (biological), ontogenentic (individual), and sociogenetic (social).  Phylogenetic adaptation is the well known adaptation of genetic material through natural selection, which also known as biological evolution.  Ontogenetic adaptation is adaptation of the individual over its lifetime and is often split into development (adaptation of morphology) and learning (adaptation of behavior).  Sociogenetic adaptation is adaption of cultural information that is communicated through social learning mechanisms.

It is clear that these three modes of adaptation will be coupled, that is, they will impact one another \citep{HN, SSE}.  We think that any acceptable model of genetic and cultural evolution must support divergence in the genetic and cultural evolutionary trajectories.  There are types of divergence possible.  If the selection pressures on the genetic information and the memetic information pull in the same direction we see divergence in the speed of evolution.  Cultural evolution is typically much quicker in this case.  Yet when the selection pressures on genetic information and cultural information pull in different directions the model should allow the genetic information and the cultural information to diverge.  

In human culture we can see that this divergence can occur in individual humans.  For instance, a Catholic priest may swear an oath of celibacy because his culture rewards him for it.  A Samurai may kill himself if he feels his cultural obligations have not been met.  Refusing to reproduce and killing oneself are both acts that are contrary to the genetic imperative but in these cases support the individual's cultural imperative.

Other more drastic cases occur when a whole culture adopts behavior contrary to the biological imperative.  This includes the celibate religious sect the Shakers from the 1770s and a number of mass suicides including  Jonestown in 1978 and the Heaven's Gate cult in 1997.  The tragic end of these cultures is usually their own destruction. 

Other models of genetic and cultural evolution have been presented but very few satisfy all of our desired properties.  In this paper we will review evolutionary models and evaluate them as models of gene-meme co-evolution using the two principles outlined above.  After reviewing relevant models we will present our model and elaborate on how it integrates benefits of multiple models and satisfies our aims.  Finally we will discuss the  potential for divergence of genetic and cultural evolutionary trajectories with our model.

\section{Prior Models}
Dawkins suggested evolution can occur in any population of information bearing agents so long as the population had three properties: heredity, variation, and selection.  This simple model of evolution we will call the \emph{basic evolutionary model}.

To demonstrate the applicability of this model to non-biological populations Dawkins coined the term \emph{meme} to name the unit of cultural selection.  He argued that populations of memes satisfy the three properties and thus also undergo a process of evolution.

Dawkins application showed that evolutionary principles can be applied to other domains.  However his model did not describe the dynamics of a system undergoing both genetic and cultural evolution.  In this section we will review the basic evolutionary model as presented by Dawkins and then review models that expand upon this model.  Our goals is to evaluate these expanded models for their value as models of gene-culture coupled systems.

\subsection{Basic Evolutionary Model}

\begin{figure}[!t]
\centering
\includegraphics[width = 240px]{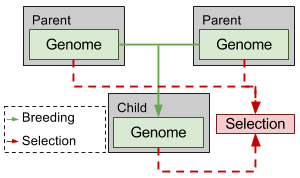}
\caption{Basic Evolutionary Model}
\label{fig:evo}
\end{figure}

Dawkins' evolutionary model is the basis of all the models we will consider in this section.  The model exists as a set of minimal conditions for a system to evolve.  The other models in this section are explorations of additional properties and mechanisms that enrich this basic model.

As mentioned above the basic evolutionary model requires a population of agents.  These agents must bear some type of information that is critical to their survival, and they must possess a means of replicating this information.  Dawkins calls these types of agents replicators.

A population of replicators is not sufficient.  The population must also have three properties.  When information is replicated it must be replicated with some degree of fidelity, that is, it must be replicated within some reasonable error rate (heredity).  The information among agents in the population must be varied (variation).  Lastly, the information among agents in the population must determine which agents are pruned from the population and which agents can replicate (selection).

At an abstract level this model makes little commitment to how replication occurs and what the rules for selection are.  This means, for instance, that both natural selection and artificial selection seem to satisfy the conditions.  Nonetheless, the standard analogy is to biology so the basic evolutionary model tends to be characterized in terms of biological evolution.

In the basic evolutionary model the information is encoded in the genome (Fig. \ref{fig:evo}).  One or more parents contribute information to a replication process that creates a new agent and genome.  In the simplest model the information in the genome is directly referenced for the selection process (i.e. there is no interpretation of the information).

Dawkins suggests that this model is the minimum required for the force of evolution to occur, not that this is a complete or proper model of biological evolution.  However researchers have implemented this basic model many times in silico demonstrating that this minimal model can lead to evolution.  Many researchers, including Dawkins, have expanded on this standard model to describe biological as well as cultural phenomena \citep{lumsden1981coevolutionary, d82, boyd1983cultural,  henrich2003evolution}.

\subsection{Agent Based Models}

\begin{figure}[!t]
\centering
\includegraphics[width = 240px]{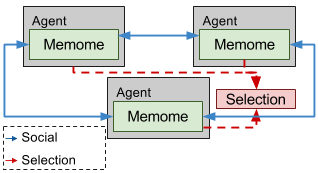}
\caption{Agent Based Model}
\label{fig:abm}
\end{figure}

Agent based models are commonly used in biology and social sciences for modeling phenomena involving populations of agents \citep{b02, e06, nh11,spss12}.  A typical agent based model consists of a population of agents.  These agents, as in the basic evolutionary model, bear information that is used to make decisions or select behavior as they interact with their environment and each other.

Information is replicated in episodes of social interaction.  Agents exchange information with each other.  Selection of behavior results in better or worse performance in the environment and so the information can evolve over time.

The information in an agent based model is now meant to represent learned information (not genetic) so we call it the \textit{memome} (Fig. \ref{fig:abm}).  While these models do not follow the standard biological characterization of the basic evolutionary model we suggest they nonetheless are explorations of variations on the basic evolutionary model. 

Agents do not die and are not born in the simplest agent based models (some do incorporate this and some kind of genetic evolution but we would classify them in one of the categories that follow).



These agent based mechanisms satisfy the three properties of the basic evolutionary model.  Information is copied with some fidelity in the replication process.  Agents clearly have different information by the design of agent based models.  Third, as argued above, selection also operates in these models.  Our conclusion is that agent based models are explorations of the parameter space of the basic evolutionary model.

The value of agent based models is unquestioned.    Nonetheless most agent based models exemplify the basic evolutionary model with a non-standard (i.e. non-biological) interpretation.

\subsection{Horizontal Transfer Model}

\begin{figure}[!t]
\centering
\includegraphics[width = 240px]{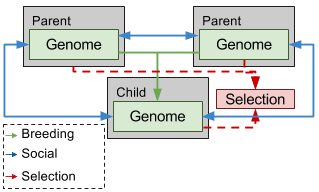}
\caption{Horizontal Transfer Model}
\label{fig:horiz}
\end{figure}

Horizontal transfer of information has been suggested as a hallmark of cultural evolution \citep{gwnb14}.  Horizontal transfer models are a blend between the basic evolutionary model and the social mechanisms used in agent based models.  The horizontal transfer model (Fig. \ref{fig:horiz}) still relies on a standard parent to child information transfer during reproduction (since this transfer is unidirectional and always passes from parents to children this is called a \emph{vertical transfer}).

The social mechanisms of agent based models also transfer information from agent to agent but these transfers are bidirectional and can typically occur anytime, not just during instances of reproduction.  In particular this means that during the lifetime of an agent it can change its internal information through these social transfers.  

This transfer is called \emph{horizontal transfer} because it can occur during the lifetime of the agent and can occur between agents of the same generation.  In particular information can be transferred in any direction including parent-to-child, child-to-parent, sibling-to-sibling and in general, agent-to-agent.

While horizontal transfer is a characteristic of cultural evolution we do not think that it is a sufficient characteristic.  It can be shown that evolutionary models with horizontal transfer have some benefits over strictly vertical transfer models \citep{thp13}.  These simulated results are also backed by research on horizontal transfer of genetic material among bacterium, plants and fungi \citep{sk01,s12}.  The biological results stress that as horizontal transfer of genetic material does occur in the natural world we should treat horizontal transfer models as models of biological evolution.  That is, the horizontal transfer model described here is a valuable enhancement of the vertical transfer interpretation in the basic evolutionary model but it does not model the gene-culture coupled system.  

\subsection{Evolutionary Developmental Model}

\begin{figure}[!t]
\centering
\includegraphics[width = 240px]{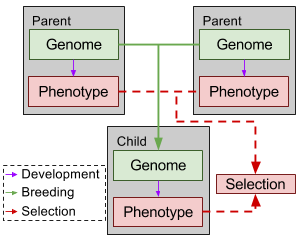}
\caption{Evolutionary and Developmental Model}
\label{fig:evodevo}
\end{figure}

Earlier we introduced the field of memetic algorithms.  The memetic algorithm model adds an additional stage to the naive evolutionary model.  Agents are bred and born as in the basic model.  However, the genetic information is not the information used for selection (as it is in the previous models).  Instead a local search is conducted around the genetic information for possibly better information.  This information is instead used for selection.

In biology this is called the genotype-phenotype distinction and the process of mapping a genotype to a phenotype is called development \citep{h12}.  Development in biology is commonly split into morphological development (growth) and behavioral development (learning).  Adding both growth and learning to evolutionary simulations has been a common improvement over the basic evolutionary models \citep{HN, SSE, MC14}.  The biological model that best captures evolution and development is called the evolutionary developmental model or evo-devo for short.  

Fig.~\ref{fig:evodevo} shows the standard evo-devo model.  The genome interacts with the environment through a process of development to produce the phenotype.  The phenotype is a second store of information that is used in selection.  However if reproduction occurs, it is only the genetic information that is passed on.

This makes a distinction between the information transferred in transfer events and the information used in evaluation for the purposes of selection.  In the basic model this was the same information.  In the evo-devo model we separate these two kinds of information into two different information stores as well as provide rules for how to develop a phenotype from a genotype.

\begin{figure}[!t]
\centering
\includegraphics[width = 240px]{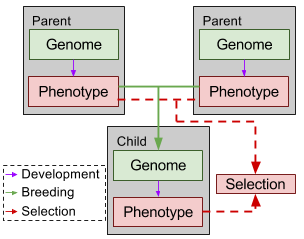}
\caption{Lamarck Model}
\label{fig:lamarck}
\end{figure}

A variant of the evo-devo model that is commonly found in memetic algorithms and other simulations corresponds to Neo-Lamarckian evolutionary theory.  The primary distinction between Neo-Lamarckian and Darwinian evolution is that learned traits can be passed on in Neo-Lamarckian models.  Fig. \ref{fig:lamarck} shows a Neo-Lamarckian evo-devo model in which the information in the phenotype is transferred during reproductive events instead of the genotype.

A Neo-Lamarckian evo-devo model can be implemented with only a single information store (genetic) in which development is internal changes to the genome through interaction with the environment.  When reproductive events occur the current state of the genome is transferred.  In these models the information added through development can be passed on in reproduction.

Evo-devo models are well studied in biology and are popular evolutionary models for simulations and optimization.  These models capture phylogenetic and ontogenetic adaptation but do not model sociogenetic adaptation.

\subsection{Evolved Social Learning Neural Networks}

The models reviewed so far have been variants or expansions of the basic evolutionary model.  Most have clear biological instances in the natural world.  While all model important processes, none succeed at modeling the interplay between genetic and cultural information.

In our opinion the best attempts to model a coupled genetic and cultural system through simulation so far have come from researchers trying to evolve neural networks that also engage in phases of social learning \citep{g95, DP, b01, s02, AP, co07, bc11}.  While these experiments have had varied levels of success we believe this type of model is on the right track.

Agents in this model have a genome that encodes an artificial neural network (typically the weights of a predetermined network topology).  Evolution of this information is carried out following the basic evolutionary model.

However, during the lifetime of the neural network the network can engage in learning.  A basic type of neural network learning is backpropagation learning.  Backpropagation training requires a supervisor and most environments are not designed to supervise learning.  In these models other networks provide the expected output to the learning network in a stage of social learning. Note that there are other possible means of training a network like neuromodulated plasticity \citep{sbmdf08}.

The primary advantage of this model is that there are two distinct information stores for genetic and cultural information.  The genome is inert and is only used to build the initial network.  The weights of the network are stored independently (from the genome) and can be seen as a second information store.  This store changes over the lifetime of the agent and is active in selecting behavior (i.e. in generating the phenotype that is relevant to selection).  

Despite properly modeling the relationship between genetic and cultural information these models suffer from drawbacks due to the choice of artificial neural networks.  Artificial neural networks still represent a very simple model of biological neural networks.  Evolving ANN weights and topology is very challenging despite new advances like the NEAT algorithm \citep{sm02}.  Transferring information from one ANN to another using supervised training is slow, unreliable, prone to errors, and artificially requires training sessions. We pick up where these experiments leave off by clearly describing the model and presenting a different implementation choice that is easier to work with than neural networks.

\section{The Dual Inheritance Model}

\begin{figure}[!t]
\centering
\includegraphics[width = 240px]{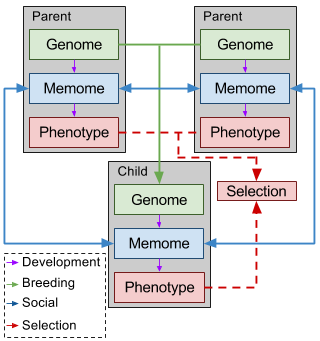}
\caption{Dual Inheritance Model}
\label{fig:dim}
\end{figure}

The dual inheritance model exploits the advantages of the evolved social learning neural networks.  Genetic and cultural information are stored separately and perform separate roles.  We pivot from neural networks and embrace evolutionary processes.  That is, we model phylogenetic, ontogentic, and sociogenetic adaptation as populations of individuals competing for fitness.

At birth a new agent inherits its genome from its parents (Fig. \ref{fig:dim}).  Through a process of development the genome produces the newborn's memome.  The genome remains inert over the lifetime of the agent while the memome is active in behavior selection and adaptive through learning.  The memome interacting with the environment creates the phenotypic information which is active in selection.  The memome is modified through interaction with the environment (learning) and through interaction with other agents (social learning).

We have implemented this model once before \citep{MC14}.  Here we describe relevant implementation details from our current system \citep{MC16a, MC16b}.  Agents in our simulation exist in a random geometric network of food sites.  During the day they spend energy moving around, foraging for food, breeding, learning and social learning.  At the end of the day their food is converted to energy.  An agent that runs out of energy dies, and one that stores enough surplus energy can reproduce.

\subsection{Genome}

The genome of an agent represents a path of sites through the random geometric network.  At each site the gene determines what strategy to use to gather food, and whether to engage in breeding, learning or social learning at that site.  The entire genome represents a very long path that cannot be completed in one day.  

During an agent's lifetime the genome plays two roles.  It must create the memome and during reproductive events it is used in recombination.  Details of our agent's genetics can be found in \citep{MC15b}.

If an agent has a surplus of energy and is prepared to breed then it must spend time breeding during the day.  It does this at a particular site at a particular time of day.  If there is another agent also performing the breed action at the same site at the same time then sexual reproduction occurs.  If not the agent must wait for its next opportunity to breed.

Upon birth the genome copies short segments of itself into the memome.  Each segment represents a path through the network long enough to be completed in a single day and starting at a specific site.  We call these segments memeplexes and we copy every possible memeplex from the genome into the memome during initial development.  We consider this technique is similar to the MAP-elites \cite{mc15} strategy for multi-objective optimization.  We want to find the best memeplex given a particular starting site so we store the best memeplex for each starting site.

\subsection{Memome}

The memome is a collection of memeplexes.  During behavior selection an agent selects the best memeplex given the current site.  First an agent gathers the memeplexes that start at this site.  The agent then selects the memeplex with the highest expected resource reward at the lowest energy cost.  This is the agent's behavior for the day.

The memome is not inert over the lifetime of an agent.  A newborn agent has memeplexes that are directly copied from the memome.  Over time new memeplexes are added to the memome through individual learning and social learning.

Individual learning in our model occurs only if the agent spends time engaged in learning during the day.  This requires the agent to select a memeplex for the day in which the agent spends time learning at at least one site.  If it does so then the memeplex will add a possibly mutated copy of itself to the memome.  This allows the agent to, among other things, optimize its foraging strategies.

Individual learning is a process in which the memome can improve itself via interaction with the environment.  This is a developmental process that is analogous to the one from the evo-devo model above.  In our experiments agents with only individual learning can improve their behavior with this mechanism but these improvements are lost when the agent dies.

Social learning is the important additional feature our agents need to achieve cumulative cultural evolution.  In order for an optimized memeplex to survive the death of its host it must be shared with another host.  Social learning in our implementation is similar to breeding.  For two agents to learn from one another they must both perform the social learning action in the same site at the same time during the day.  If they do they exchange a possibly mutated copy of the memeplex that they used that day.  We treat this exchange as roughly approximating the agents telling each other what they did that day.

Individual learning is the power that allows the agent to optimize its behavior.  This optimization is lost if it is not shared.  Social learning is the power that allows a population to share optimized behaviors.  The collection of all agents' memeplexes is called the memosphere.

A memeplex's evolutionary goal is to maintain copies of itself in the memosphere.  To do this it must be optimized and so this usually means it spends time learning.  Further more it is critical that it spend time social learning so that it can spread itself to other agents.  Note that a memeplex is not concerned with spending time breeding.  Breeding does not help the memeplex spread or optimize.  This will be a source of divergence.

\section{Divergence}

\begin{figure}[!t]
\centering
\includegraphics[width = 240px]{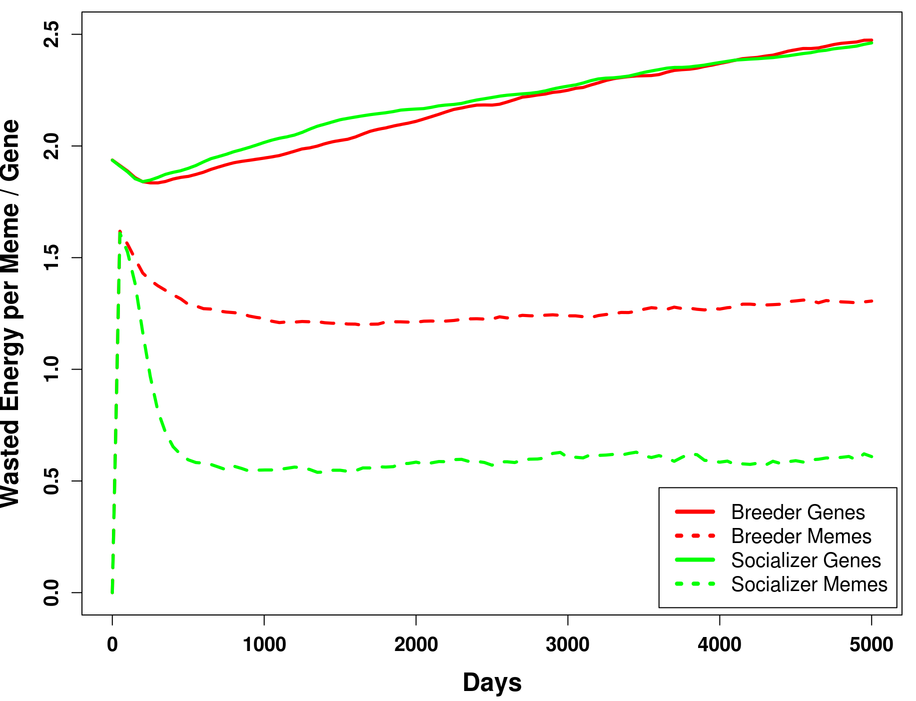}
\caption{Contrasting the average genetic optimization and average memeplex optimization over time for two population of agents.  Breeders are a control group using the evo-devo model.  Socializers follow the dual inheritance model.  }
\label{fig:opt}
\end{figure}

We have observed divergence of trajectories in both our prior work and in the model described here.  There are two types of divergence that we have observed.  First both the genome and the memeplexes are trying to optimize themselves.  The memeplexes have many opportunities to improve while the genome only has an opportunity when it reproduces.  This means that the memeplexes will optimize more rapidly than the genome.  Both trajectories are in the same direction but one gets there much faster.  We call this divergence under cooperative selection pressures.  Fig.~\ref{fig:opt} summarizes results from our current implementation \citep{MC16a}.

The second type of divergence occurs when the selection pressures are competitive.  We mentioned above that breeding does not help the memeplex.  In fact it hinders it as time is better spent foraging than breeding and the memeplex needs to be as optimal as possible.  Selection pressure against breeding in the memosphere is strong and we see many optimized memeplexes that spend no time breeding.  Contrary to this selection pressure for breeding is very strong in the genome.  Most agents have genomes that allocate a lot of time to breeding.  This is a case of divergence under competitive selection pressures.

This divergence has an interesting effect.  Memeplexes that spend no time breeding will suppress breeding in the agents that select them.  This can create a culture of celibacy.  If this culture becomes dominant it runs the risk of wiping out the population.  We do indeed observe this in our current implementation.  21 out of 100 runs ended with complete population collapse before 5000 days.  This was not observed in our prior work.  In \citep{MC14} agents always have the chance to reproduce and so colony collapse was not possible for this reason.

\begin{figure}[!t]
\centering
\includegraphics[width = 240px]{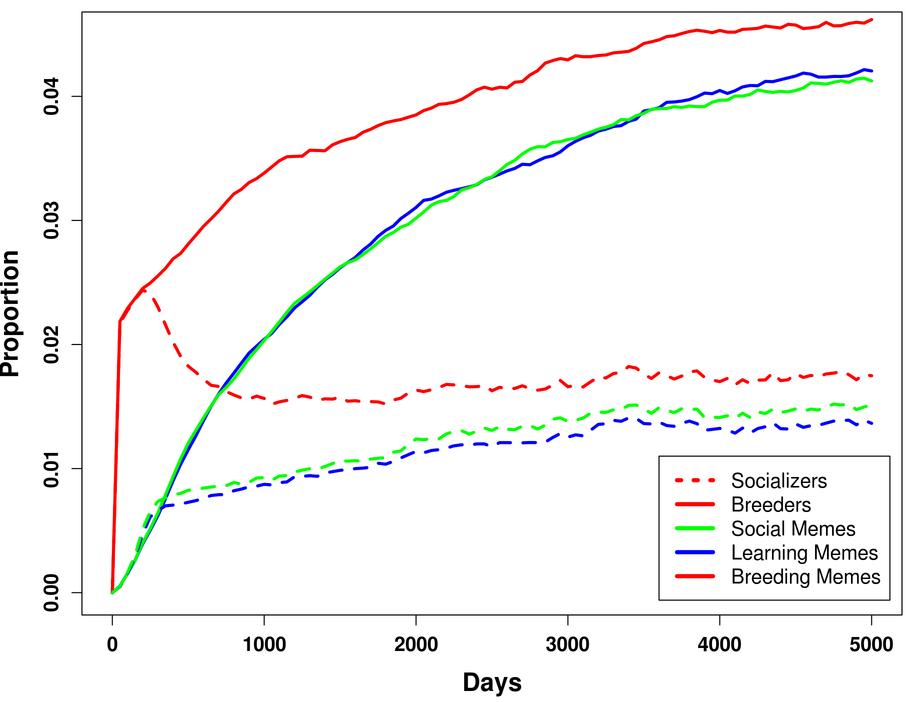}
\caption{Contrasting the average time spent breeding, learning and socializing between two populations of agents.  Breeders are a control group using the evo-devo model.  Socializers follow the dual inheritance model.}
\label{fig:memes}
\end{figure}

This divergence also occurs in both implementations relative to learning and social learning.  We observe that there is strong selection pressure for learning and social learning in the genome.  This is due to the benefits these adaptive mechanisms grant the agent from an evolutionary perspective.  While there is selection pressure for learning and social learning in the memosphere there is also selection pressure to optimize these processes as much as possible.  This means that once learning no longer pays off it is also commonly eliminated.  Social learning is commonly minimized as much as possible to ensure a very optimal memeplex that can still spread itself.   Fig.~\ref{fig:memes} summarizes results from our current implementation \citep{MC16a}.

This divergence leads to a number of phenomena we wish to explore in more detail.  Young agents have behaviors dictated largely by their genome.  This means they spend a lot of time breeding, learning, and social learning.  In a mature culture at some point the young agent will learn optimized memeplexes from others and alter its behavior.  It will often no longer spend time breeding or learning.  In any case it will spend as little time as possible breeding, learning and social learning.  When it engages in social learning it usually does not get a memeplex better than its current ones, instead it is sharing its collection of memeplexes with others.

This causes a phenomena where young agents breed, learn and social learn more often than older agents.  We observe that this is the case also in humans.  Young humans engage in a considerable amount of learning and social learning.  Young adults are more likely to have children than older adults.  We hope to explore if this is an artifact of our implementations or a feature of other dual inheritance models.

\section{Discussion}

We have incorporated the benefits of all of the discussed models into our own with an attempt to capture a model that includes both genetic and memetic co-evolution as accurately as possible.  Our model has the vertical transfer of the biological model as well as the horizontal transfer of the cultural model while avoiding the drawbacks of a single information storage that previous models suffer.

The key benefit of additional information storage is that the two stores can diverge.  This idea is already present in the  evo-devo model.  In this model the information of the genotype is passed on through reproduction while the information in the phenotype is used for selection.  The benefit of this model (in both biology and simulation) is that the information in the phenotype can adapt over the lifetime to benefit passing on the information in the genotype of an agent.

This has two-fold benefit.  First, the phenotype is free to adapt to any circumstance facing the agent during its lifetime.  Second, the genotype is not disturbed in this adaptation and can be passed on intact to the next generation.

We can consider this benefit as a divergence of adaptive trajectories between the genotype and the phenotype.  The genotype can still focus on replicating itself while the phenotype can focus on keeping the agent alive.

One advantage to the genotype in this arrangement is that the phenotype has a limited lifespan.  After the agent dies the adaptations in the phenotype are lost and cannot upset the genotype's goal of replication.  In this case, the phenotype is subordinate to the genotype and can only adapt within the boundaries dictated by the genotype.  While these information stores can diverge from one another we say the phenotype in these models is \emph{tethered} to the genotype.  It can diverge but no further than the genotype allows \citep{MC14}.

Our model adds a new layer of information: the memome.  The memome is also distinct from the genome and so can evolve on its own trajectory.  Thanks to social learning it can avoid the death sentence of the phenotype.  When the agent dies, if it has spread its cultural information to another agent, its memes can still live on.  Unlike the information in the phenotype that exists tethered to the genotype, the memotype is free to evolve along its own trajectory.

The information in the memome is not completely free of the genome.  In human culture, and in our simulations, the existence of memes is still dependent upon the existence of the agents that house them.  These agents are biological and thus if the memome diverges so far as to endanger the genome it may endanger itself as well. So when we discuss divergence we do not mean a complete decoupling of genetic and cultural systems, but rather a very long leash.  Cultures are able to destroy their host but in so doing they destroy themselves as well.

It is worth noting that our model does not expand on the implicit model of evolved social learning neural networks mentioned above.  The implicit model in these experiments is identical to our model.  One improvement we have made in terms of implementation is to make better choices of underlying structures and mechanisms.  In particular, for simplicity, we have modeled the genome and memome as storing the same type of information.  As a result our learning and social learning processes mimic the underlying genetic mechanisms of mutation and recombination.  This makes modeling and implementing these processes much simpler and we believe this contributes to the success of our agents.

\section{Acknowledgments}

Jobran Chebib was supported by the Swiss National Science Foundation (grant PP00P3\_144846/1 awarded to Fr\'ed\'eric Guillaume).  The authors would like to thank James Borg and anonymous reviewers for helpful comments on an early draft.

\footnotesize
\bibliographystyle{apalike}
\bibliography{example}

\end{document}